\documentclass[10pt,twocolumn,letterpaper]{article}

\newcommand{\projectpage}{\url{https://gc-motion.pages.dev}}

\usepackage[pagenumbers]{cvpr} %

\usepackage{multirow}
\usepackage{makecell}
\usepackage{amsthm}
\theoremstyle{definition}
\newtheorem{definition}{Definition}
\theoremstyle{plain}
\newtheorem{proposition}{Proposition}
\newcommand{\ours}{GC-Motion\xspace}
\makeatletter
\newbox\gc@labels
\let\gc@shipouthook\@EveryShipout@Hook
\def\@EveryShipout@Hook{%
  \global\setbox\gc@labels\box\@labels
  \if@inlabel\gdef\gc@inlabel{\global\@inlabeltrue}\else\gdef\gc@inlabel{}\fi
  \global\@inlabelfalse
  \gc@shipouthook
  \global\setbox\@labels\box\gc@labels
  \gc@inlabel}
\makeatother

\newcommand{\lleg}{\lambda_\text{leg}}

\newcommand{\second}[1]{\underline{#1}}
\usepackage{colortbl}
\definecolor{oursbg}{HTML}{E4F1F0}
\newcommand{\oursrow}{\rowcolor{oursbg}}

\definecolor{cvprblue}{rgb}{0.21,0.49,0.74}
\usepackage[pagebackref,breaklinks,colorlinks,allcolors=cvprblue]{hyperref}
\usepackage[capitalize]{cleveref}

\def\confName{CVPR}
\def\confYear{2027}

\title{GC-Motion: Streaming Text-to-Motion with Granularity-Aware\\Conditioning and Contact-Guided Decoding}

\author{Leo Fung \qquad Jamie \qquad Jack Yu \qquad Zhao Wang$^{\dagger *}$\\[3pt]
LimX Dynamics\\[3pt]
{\small $^\dagger$Project lead \qquad $^*$Corresponding author}\\[2pt]
{\small Project page: \projectpage}
}

\begin{document}
\twocolumn[{%
\renewcommand\twocolumn[1][]{#1}%
\maketitle
\begin{center}
\captionsetup{type=figure}
\includegraphics[width=\textwidth]{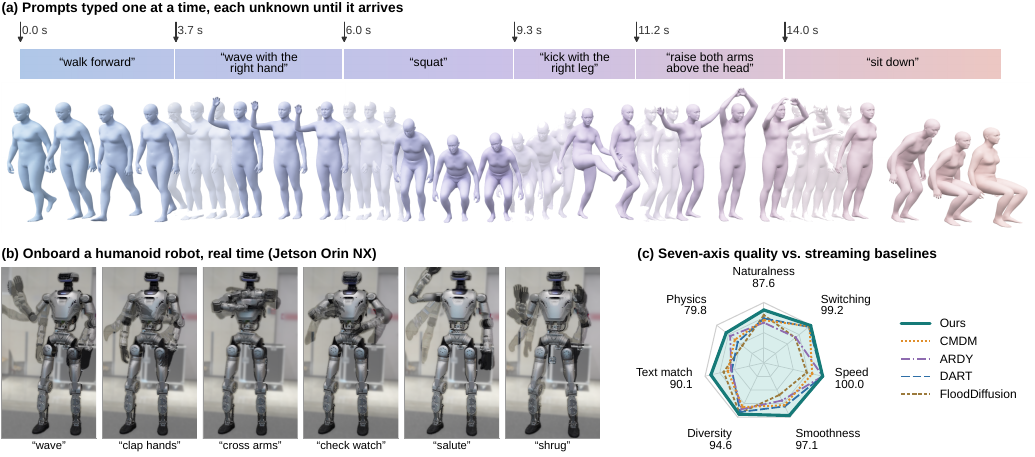}
\caption{\ours: streaming text-to-motion.
(a)~Prompts arrive one at a time, each unknown in advance; the generated motion stays continuous across them.
(b)~The same model onboard a humanoid robot in real time, with causal foot locking.
(c)~Seven-axis scores (0--100) against streaming baselines (\cref{tab:main}).}
\label{fig:teaser}
\end{center}
\vspace{0.5em}
}]
\begin{abstract}
Streaming text-to-motion generates continuous motion from instructions that arrive one at a time, without access to future prompts. Training such a model should assign each piece of text to the motion it describes, yet many motion datasets caption entire clips without marking where each action begins and ends. We present \ours, a streaming text-to-motion framework that learns from annotations of different temporal granularity. Its granularity-aware conditioning uses temporally aligned descriptions to supervise token-level diffusion forcing, while clip- and class-level captions train the same generator through whole-window denoising. A generated motion can still exhibit foot sliding when root displacement and local pose are decoded separately. Our contact-guided decoder uses predicted foot contacts and foot kinematics to adjust the horizontal root displacement before each block is emitted. Experiments show that granularity-aware conditioning reduces heading drift when incorporating additional motion data with mixed annotations. With the generator and latents fixed, contact-guided decoding lowers median stance-ankle speed from 5.08 to 1.44 cm/s. Compared with four streaming baselines, \ours scores higher on text matching, smoothness, and physics while meeting our real-time targets. We further demonstrate real-time onboard generation on a full-size humanoid robot.
\end{abstract}

\section{Introduction}
\label{sec:intro}

A user tells a humanoid robot ``walk forward'' and, a few seconds later, ``turn left''.
While the robot walks, it knows neither what the next instruction will say nor when it will arrive.
The motion must start within a fraction of a second, continue at a fixed rate, and change when the instruction arrives, without revising frames that have already been committed.
This differs from synthesizing a clip from one caption~\cite{guo2022humanml3d,tevet2023mdm,zhang2023t2mgpt,guo2024momask}.
Streaming generators meet these timing constraints with autoregressive rollouts or diffusion forcing~\cite{zhao2025dart,xiao2025motionstreamer,cai2025flood,zhao2026ardy,yu2026cmdm,chen2024diffusionforcing}.
Beyond timing, this raises two questions.
During training, which frames should each caption supervise?
During decoding, how can stance feet stay planted in blocks that are committed one at a time?

The first question concerns the training text.
Text aligned with the stream, such as the frame-level labels of BABEL~\cite{punnakkal2021babel}, is scarce; most motion data are captioned clip by clip for offline generation, yet they add a much wider range of motion.
A caption such as ``walk forward, turn left, then stop'' tells which actions happen and in which order, but not where one ends and the next begins.
Copied to every streaming token, it makes the tokens of the straight walk read ``turn left'' as well, and a model trained this way drifts when told to walk straight (\cref{tab:offline}).
A streaming generator must learn from such data without treating them as aligned.

The second question concerns foot contact in the committed blocks.
During walking, the pelvis moves forward while the stance foot moves backward relative to it, and the two cancel in the world.
Decoders that output the root displacement and the local pose as separate channels do not enforce this cancellation, so the model itself makes feet slide.
A codec round trip alone raises the median stance-ankle speed of motion capture from 0.93 to 6.70\,cm/s (\cref{sec:output}).
Post-processing can repair the output, but not by looking ahead, because a stream commits each block once it is decoded.

\begin{figure*}[t]
\centering
\includegraphics[width=\textwidth]{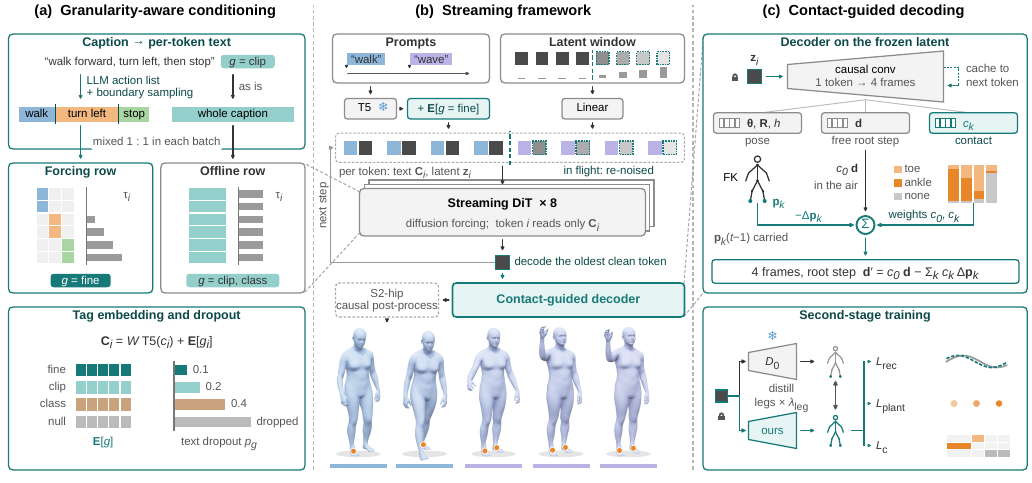}
\caption{Overview of \ours. (a)~Granularity-aware conditioning (\cref{sec:offline}): planned fine text trains forcing rows token by token; clip- and class-level captions train offline rows only. (b)~Streaming framework: at inference, each step denoises a window in which every token reads only the prompt active at its time; once the oldest token reaches zero noise, the decoder converts it into four frames and commits them to the stream. (c)~Contact-guided decoding (\cref{sec:contact}): the root displacement is composed from predicted contacts and foot kinematics, so the root step reduces stance-foot sliding.}
\label{fig:method}
\end{figure*}

We present \ours, which answers each question where it arises (\cref{fig:method}).
Granularity-aware conditioning changes which training rows a caption supervises, and contact-guided decoding changes how the root displacement is computed when latents become frames.
Our contributions are:
\begin{itemize}[leftmargin=*,itemsep=1pt,topsep=2pt]
  \item We introduce \textbf{granularity-aware conditioning} for streaming text-to-motion, which routes each caption by the time span it describes, so that aligned descriptions condition individual tokens in diffusion forcing and clip- and class-level captions train the same generator through whole-window denoising.
  \item We introduce a \textbf{causal contact-guided decoder} that combines a predicted root displacement with contact-point kinematics per frame. It reduces stance-foot sliding in raw output and keeps the latent of an existing generator.
\end{itemize}
Granularity-aware conditioning reduces the heading drift observed when coarse captions are copied across tokens and keeps the coverage that the added data bring; with the generator and latents fixed, contact-guided decoding lowers the raw stance-ankle speed from 5.08 to 1.44\,cm/s.
Compared with four streaming baselines, the complete system scores higher on text matching, smoothness, and physics while meeting our real-time targets (\cref{tab:main}).
We also report the standard HumanML3D protocol with a HumanML3D-only generator and run onboard a full-size humanoid robot with causal foot locking (\cref{fig:teaser}).

\section{Related Work}
\label{sec:related}

\paragraph{Streaming text-to-motion.}
Offline models generate a clip from one caption~\cite{tevet2023mdm,chen2023mld,zhang2023t2mgpt,guo2024momask} or chain captions on a timeline given in advance~\cite{athanasiou2022teach,shafir2024priormdm,barquero2024flowmdm}.
Streaming models instead receive each instruction without knowing in advance what it says or when it arrives. DART~\cite{zhao2025dart} rolls out latent motion primitives, MotionStreamer~\cite{xiao2025motionstreamer} predicts continuous causal latents, ARDY~\cite{zhao2026ardy} uses autoregressive diffusion, and CMDM~\cite{yu2026cmdm} pairs a causal VAE with diffusion forcing.
FloodDiffusion~\cite{cai2025flood} provides the framework we build on, a rolling window denoised with a triangular diffusion-forcing schedule~\cite{chen2024diffusionforcing} and time-varying text; FloodDiffusion~2~\cite{cai2026flooddiffusion2} makes it cheaper and adds path control.
Streaming video models pipeline denoising in the same way~\cite{kim2024fifo,yin2025causvid,huang2025selfforcing,yang2025longlive,magi1}.
\ours keeps this framework and changes two things, which time span each caption supervises during training and how each latent is decoded causally into frames.

\paragraph{Text supervision at different time scales.}
HumanML3D~\cite{guo2022humanml3d} captions whole clips and BABEL~\cite{punnakkal2021babel} provides frame-level action labels; larger capture collections are mostly captioned clip by clip~\cite{lin2023motionx,ji2026hiphi}.
Methods that compose actions in time take segment-level text, learned from BABEL~\cite{athanasiou2022teach} or placed on a user-given timeline at test time~\cite{petrovich2024stmc}.
In video, narrations are only weakly aligned with frames~\cite{miech2019howto100m}, and alignment networks learn where each sentence applies~\cite{han2022tan}.
Closest to our setting, UniMotion~\cite{li2025unimotion} conditions generation on global and frame-level text, and SegMo~\cite{dang2026segmo} aligns text phrases with motion segments contrastively.
We instead let the time span of a caption decide how it trains the generator, through token-level diffusion forcing or through whole-window denoising, and split only multi-action captions, by sampled plan boundaries.

\paragraph{Foot contact and humanoid execution.}
Latent and token-based generators~\cite{chen2023mld,zhang2023t2mgpt,guo2024momask,cai2025flood} decode the root velocity as a channel next to the joint rotations, untied from the stance feet.
Motion features often include binary foot contacts that mark when a foot is down~\cite{guo2022humanml3d}, and MDM penalizes the velocity of feet labeled as in contact~\cite{tevet2023mdm}, but both encourage rather than enforce planted feet.
Keyframe synthesis can reconstruct the root trajectory from predicted stance feet and posture~\cite{li2024footconstrained}, and PhysDiff projects denoising steps through a physics simulator~\cite{yuan2023physdiff}.
Streaming systems usually remove sliding after generation, by foot locking and contact-aware inverse kinematics, by ARDY's foot-skate correction~\cite{zhao2026ardy}, or by physics-based tracking~\cite{tevet2025closd}, often over windows that include future frames.
Our decoder instead composes the root displacement from per-frame contact weights and foot kinematics, inside a causal decoder on a frozen latent.
For humanoids, text-driven control streams generated motion to a tracking policy~\cite{xie2026textop,mao2024uh1,shao2025langwbc}, and whole-body tracking~\cite{he2024omnih2o,cheng2024exbody} executes retargeted~\cite{araujo2025gmr} human motion; we stream into such a pipeline, with causal foot locking before the tracker.

\section{Method}
\label{sec:method}
\label{sec:analysis}
We define the task and backbone (\cref{sec:backbone}, \cref{fig:method}(b)). Granularity-aware conditioning (\cref{sec:offline}, \cref{fig:method}(a)) decides which frames each caption supervises, and contact-guided decoding (\cref{sec:contact}, \cref{fig:method}(c)) how each fully denoised token is decoded into the frames that are committed; each starts from the mismatch it addresses, and \cref{sec:train} covers training and inference.

\subsection{Streaming formulation and backbone}
\label{sec:backbone}
\begin{definition}[Streaming text-to-motion]
\label{def:stream}
Instructions arrive as a stream $\mathcal{I}=\{(c_j,t_j)\}_{j\ge1}$ with $t_1<t_2<\cdots$; the instruction active at time $t$ is $c(t)=c_j$ for $t\in[t_j,t_{j+1})$.
A streaming generator emits motion $\mathbf{x}_0,\mathbf{x}_1,\dots$ at 30\,fps in \emph{commits} of four frames.
Commit $i$, emitted at wall-clock time $s_i$, may depend only on $\{(c_j,t_j): t_j\le s_i\}$ and on $\mathbf{x}_{<4i}$, and is never revised.
\end{definition}
We require a first-frame latency below 1\,s, at least twice real-time throughput, and a median \emph{command latency} below 100\,ms, measured from $t_j$ to the first committed frame conditioned on $c_j$.

Frame $\mathbf{x}_t$ holds the 6D rotations $\boldsymbol{\theta}_t$ of 21 joints, the root orientation in a camera and a gravity view~\cite{shen2024gvhmr}, the root velocity, and the root height; the velocity gives the horizontal root displacement $\mathbf{d}_t\in\mathbb{R}^2$.
A causal convolutional VAE maps four frames to one 16-d token and back,
\begin{equation}
  \mathbf{z}_i=\mathcal{E}(\mathbf{x}_{4i:4i+3}),\qquad \hat{\mathbf{x}}_{4i:4i+3}=\mathcal{D}(\mathbf{z}_{\le i}),
  \label{eq:codec}
\end{equation}
so one token is one commit (133\,ms) and one action spans many tokens.
The generator follows~\cite{chen2024diffusionforcing,cai2025flood}: a Wan-style~\cite{wan2025} DiT~\cite{peebles2023dit} (8 blocks, width 1024, 111\,M parameters) over a window of $N{=}30$ tokens (4\,s), each cross-attending to the T5~\cite{raffel2020t5} features of its caption.

\paragraph{Rolling-window diffusion forcing.}
\label{sec:prelim}
Every token in the window carries its own noise level $\tau_i\in[0,1]$,
\begin{equation}
  \mathbf{z}_i^{\tau_i}=(1-\tau_i)\,\mathbf{z}_i+\tau_i\,\boldsymbol{\epsilon}_i,\qquad \boldsymbol{\epsilon}_i\sim\mathcal{N}(\mathbf{0},\mathbf{I}),
  \label{eq:interp}
\end{equation}
and the DiT predicts the flow velocity $\hat{\mathbf{v}}=v_\Theta(\mathbf{Z}^{\boldsymbol{\tau}},\boldsymbol{\tau},\mathbf{C})$ of all tokens, with per-token text features $\mathbf{C}=(\mathbf{C}_1,\dots,\mathbf{C}_N)$.
With the implied clean latent $\hat{\mathbf{z}}_i=\mathbf{z}_i^{\tau_i}-\tau_i\hat{\mathbf{v}}_i$, the denoising loss is
\begin{equation}
  \mathcal{L}_\text{DF}(\mathbf{Z},\boldsymbol{\tau},\mathbf{C})=\textstyle\sum_{i=1}^{N}\lVert\hat{\mathbf{z}}_i-\mathbf{z}_i\rVert_2^2 .
  \label{eq:df}
\end{equation}
Streaming uses the triangular schedule of~\cite{cai2025flood}, in which the noise level rises with the token index,
\begin{equation}
  \tau_i(\phi)=\mathrm{clip}\big(1+i/n_s-\phi,\;0,\;1\big),
  \label{eq:tri}
\end{equation}
with ramp $n_s{=}5$ and a phase $\phi$; training samples $\phi\sim\mathcal{U}(0,N/n_s)$.
At inference, every network evaluation advances $\phi$ by $\Delta\phi{=}0.1$ and takes the Euler step
\begin{equation}
  \mathbf{z}_i\leftarrow\mathbf{z}_i+\big(\tau_i(\phi+\Delta\phi)-\tau_i(\phi)\big)\,\hat{\mathbf{v}}_i ,
  \label{eq:euler}
\end{equation}
so each token receives 10 steps and the oldest token reaches $\tau{=}0$ every two evaluations; it is then decoded by $\mathcal{D}$ of \cref{eq:codec} into four frames and committed.

\begin{figure}[t]
\centering
\includegraphics[width=\columnwidth]{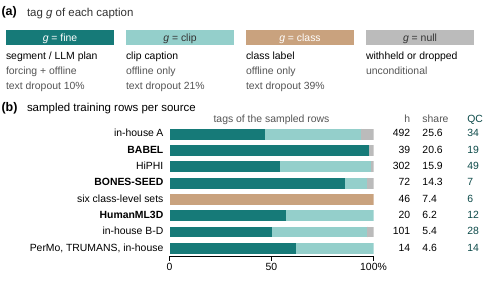}
\caption{Granularity-aware conditioning. (a)~The tag of a caption decides the rows it trains and its text dropout (realized rates). (b)~Tags of the sampled training rows per source, sorted by share (\%); about half of every source except the class-level sets is drawn as forcing rows. h: hours before QC (bold: the 131\,h base set); QC: \% of hours removed by quality control.}
\label{fig:reading}
\end{figure}

\subsection{Granularity-aware conditioning}
\label{sec:offline}
\paragraph{Supervision mismatch.}
\label{sec:input}
To train on a clip whose caption $c$ covers frames $[a,b)$, the caption must be turned into per-token conditions $c_i$.
Copying $c$ to every token makes the generator fit, for each token,
\begin{equation}
  p_\text{train}(\mathbf{z}_i\mid c)=\textstyle\sum_{u} p(u\mid c)\,p(\mathbf{z}_i\mid u,c),
  \label{eq:mixture}
\end{equation}
where $u$ ranges over the local actions inside the clips captioned $c$.
For an aligned label (a BABEL segment, or a caption of one sustained action), $u$ is fixed by $c$.
A caption such as ``walk forward, turn left, then stop'' marks no action boundaries, so copying it to every token also assigns ``turn left'' and ``stop'' to the tokens of the straight walk.
Whole-clip captions are the norm; the largest added source, a third batch of in-house capture and HiPHI~\cite{ji2026hiphi} (\cref{fig:reading}(b)), has one caption set per take of 160 to 1,900 tokens (median per source).

Copying a clip-level caption to every token thus gives some tokens text that describes other parts of the clip. We therefore use the scope of each caption to determine how it supervises the generator.

\begin{definition}[Caption granularity]
\label{def:gran}
An annotation $(c,[a,b),g)$ pairs a caption $c$ with the frames it describes and a tag $g\in\{\text{fine},\text{clip},\text{class},\text{null}\}$.
It is \emph{fine} if $c$ describes every token in $[a,b)$, \emph{clip} if it describes the frames only jointly or vaguely, \emph{class} if it names a category, and \emph{null} if no text is used, because it is withheld from forcing rows or dropped for guidance.
\end{definition}
Segment labels and captions of single- or repeated-action clips are fine (\cref{fig:reading}).
By the LLM's labels they cover 62\% of HumanML3D and 84--100\% of the third-batch sources, so routing changes only how multi-action, vague, and class-level captions are used.
For a clip-level caption that names $K$ actions in order, we derive a fine \emph{plan}: an LLM (Qwen3-235B) extracts the actions $(u_1,\dots,u_K)$ from the caption set, and boundaries $\mathbf{b}=(b_1<\dots<b_{K-1})$ are sampled each time the clip is loaded from
\begin{equation}
  p(\mathbf{b}\mid\mathbf{x},u_{1:K})\propto\exp\Big(\beta\textstyle\sum_{k=1}^{K}\sum_{t=b_{k-1}}^{b_k-1}S_{k,t}\Big),
  \label{eq:plan}
\end{equation}
where $S_{k,t}$ is a text--motion retrieval score of action $u_k$ for the window at frame $t$ (from a retrieval model separate from the evaluator), $\beta{=}2$, $b_0{=}a$, $b_K{=}b$, and every segment lasts at least 0.5\,s.
We sample $\mathbf{b}$ exactly by forward filtering and backward sampling and move each boundary 0.3\,s earlier, so that text never lags motion.

\paragraph{Two kinds of rows.}
A training window becomes a \emph{forcing row} or an \emph{offline row}.
A forcing row uses the triangular schedule and gives token $i$ the fine text of its frames, $c_i=\ell(4i)$, where $\ell(f)$ is the text that a fine annotation or plan assigns to frame $f$.
An offline row puts all tokens at one noise level $\tau\sim\mathcal{U}(0,1)$ and gives every token the same caption, which is all that a clip-level caption supports.
With $\mathcal{A}_\text{fine}$ and $\mathcal{A}$ the fine and all annotations, the generator minimizes
\begin{equation}
\begin{split}
  \mathcal{L}_\text{gen}={}&\tfrac12\,\mathbb{E}_{\mathcal{A}_\text{fine},\phi}\,\mathcal{L}_\text{DF}\big(\mathbf{Z},\boldsymbol{\tau}(\phi),\mathbf{C}^\text{tok}\big)\\
  &+\tfrac12\,\mathbb{E}_{\mathcal{A},\tau}\,\mathcal{L}_\text{DF}\big(\mathbf{Z},\tau\mathbf{1},\mathbf{C}^\text{win}\big),
\end{split}
  \label{eq:gen}
\end{equation}
\ie, batches mix the two row types 50/50.
A clip- or class-level caption is thus a target only of whole-window denoising, and the text of every forcing-row token comes from an aligned annotation or a sampled plan; how far this removes the mixture of \cref{eq:mixture} depends on the plan boundaries of \cref{eq:plan}.

\paragraph{Tag embedding and dropout.}
The text features of token $i$ are $\mathbf{C}_i=W\,\mathrm{T5}(c_i)+\mathbf{E}[g_i]$ with a learned tag embedding $\mathbf{E}$ initialized to zero.
For classifier-free guidance~\cite{ho2022cfg}, the text is dropped with a tag-dependent probability $p_g$ ($g{=}\text{null}$ marks a dropped condition): 0.1 on fine rows and 0.2 and 0.4 on clip and class rows (realized: 10, 21, and 39\%), so coarse text steers the model less.
At inference, every token reads the instruction that has arrived for its time, with $g{=}\text{fine}$ and guidance scale $w{=}2$:
\begin{equation}
  \tilde{\mathbf{v}}=v_\Theta(\cdot,\varnothing)+w\big(v_\Theta(\cdot,\mathbf{C}\mid g{=}\text{fine})-v_\Theta(\cdot,\varnothing)\big).
  \label{eq:cfg}
\end{equation}

Data preparation, \ie, window-level quality control (removing 362\,h, 33.3\% of the 1,086\,h), action balancing, and caption rewriting for straight walking, is not part of the method.

\begin{figure}[t]
\centering
\includegraphics[width=\columnwidth]{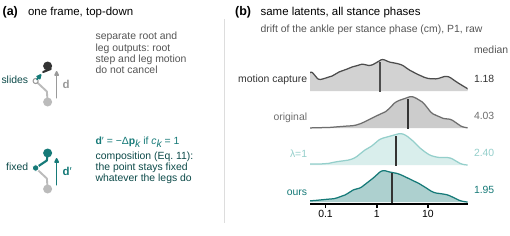}
\caption{Contact-guided decoding. (a)~With separate root and leg outputs, the root step and the leg motion of a planted foot do not cancel; \cref{eq:compose} moves the root against the contact point. (b)~The latents of the final generator on all 1,319 P1 timelines decoded by the original decoder, $\lleg{=}1$, and ours: horizontal drift of the ankle over each stance phase, against motion capture on the same segments (bar: median).}
\label{fig:committing}
\end{figure}

\subsection{Contact-guided decoding}
\label{sec:contact}
\paragraph{Root--pose mismatch.}
\label{sec:output}
Let $\mathbf{p}_{k,t}\in\mathbb{R}^2$ be the horizontal position of foot point $k$ relative to the pelvis in world orientation, so its world position is $\mathbf{X}_{k,t}=\mathbf{r}_t+\mathbf{p}_{k,t}$ for root position $\mathbf{r}_t$, and per frame
\begin{equation}
  \Delta\mathbf{X}_{k,t}=\mathbf{d}_t+\Delta\mathbf{p}_{k,t},\qquad \Delta\mathbf{p}_{k,t}=\mathbf{p}_{k,t}-\mathbf{p}_{k,t-1}.
  \label{eq:slide}
\end{equation}
A planted point requires $\mathbf{d}_t=-\Delta\mathbf{p}_{k,t}$; during walking, both correspond to speeds of about 1\,m/s and must cancel.
A decoder that outputs $\mathbf{d}_t$ and $\boldsymbol{\theta}_t$ (hence $\mathbf{p}_{k,t}$) as separate channels does not tie them, so an error in either appears as sliding.
Bypassing the generator, a codec round trip of motion capture raises the median stance-ankle speed on BABEL validation from 0.93 to 6.70\,cm/s.
Replacing either channel alone by ground truth makes it worse (7.67 and 9.84 \vs 6.25\,cm/s on a 288-clip subset), so the error lies in the pairing.
As a stream cannot use future frames, our decoder ties the two channels where tokens become frames (\cref{fig:committing}(a)).

From the frozen latent it predicts the pose, a free horizontal root displacement $\mathbf{d}_t$, and per-frame contact probabilities $c_{0,t}$ (no contact) and $c_{k,t}$ for the candidate points $k\in\mathcal{K}$, the left and right ankle and toe, with $c_{0,t}+\sum_k c_{k,t}=1$.
Forward kinematics of the decoded pose gives $\mathbf{p}_{k,t}$, and the decoder outputs
\begin{equation}
  \mathbf{d}'_t=c_{0,t}\,\mathbf{d}_t-\textstyle\sum_{k\in\mathcal{K}}c_{k,t}\,\Delta\mathbf{p}_{k,t},
  \label{eq:compose}
\end{equation}
keeping its absolute root height.
Without contact ($c_{0,t}{=}1$), the predicted displacement is kept, $\mathbf{d}'_t=\mathbf{d}_t$.
With all weight on one point ($c_{k,t}{=}1$), $\mathbf{d}'_t=-\Delta\mathbf{p}_{k,t}$, so by \cref{eq:slide} that point does not move in the world.
Several contact points share one root displacement, so a single point can still slide when the points move relative to each other:

\begin{proposition}[Residual sliding]
\label{prop:slide}
Let $\mathbf{q}_t=\sum_k w_{k,t}\Delta\mathbf{p}_{k,t}$ with $w_{k,t}=c_{k,t}/(1-c_{0,t})$ for $c_{0,t}<1$ be the contact-weighted displacement. Under \cref{eq:compose}, the world displacement of any point $k$ is
\begin{equation}
  \Delta\mathbf{X}_{k,t}=c_{0,t}\big(\mathbf{d}_t+\mathbf{q}_t\big)+\big(\Delta\mathbf{p}_{k,t}-\mathbf{q}_t\big).
  \label{eq:residual}
\end{equation}
\end{proposition}
The proof substitutes \cref{eq:compose} into \cref{eq:slide}.
The first term vanishes under confident contact, whatever the errors in $\mathbf{d}_t$ or the legs; the second, the motion of point $k$ relative to the weighted contact points (heel against toe, or foot against foot in double support), depends only on the pose and the contact weights, so the root cannot remove it.
\Cref{eq:compose} needs only frames $t$ and $t{-}1$; the last frame's state is carried across four-frame blocks, so decoding is causal and block-wise decoding matches full-sequence decoding.

\paragraph{Training on a frozen latent.}
With the encoder and latent frozen, the decoder is trained as a second stage and can replace the decoder of a trained generator.
Starting from the original decoder $\mathcal{D}_0$, it minimizes
\begin{equation}
  \mathcal{L}_\text{dec}=\mathcal{L}_\text{rec}+10\,\textstyle\sum_{m}\omega_m\lVert\mathbf{y}_m-\mathbf{y}^{0}_m\rVert_1+\lambda_\text{p}\,\mathcal{L}_\text{plant}+\mathcal{L}_\text{c},
  \label{eq:dec}
\end{equation}
where $\mathcal{L}_\text{rec}$ is the reconstruction loss on the composed trajectory, the second term distills every channel $m$ but the root velocity to $\mathcal{D}_0$ ($\omega_m{=}\lleg$ for leg joints, 1 otherwise), $\mathcal{L}_\text{plant}=\sum_{t,k}\mathbb{1}[k\text{ planted}]\lVert\Delta\mathbf{X}_{k,t}\rVert$, and $\mathcal{L}_\text{c}$ is a contact cross-entropy against soft labels; we fine-tune for 15k steps.
Full leg distillation ($\lleg{=}1$, $\lambda_\text{p}{=}0.5$) keeps the leg pose of $\mathcal{D}_0$ and hence the second term of \cref{eq:residual}.
Our default lowers the leg distillation weight to $\lleg{=}0.1$ and raises the planted-foot weight to $\lambda_\text{p}{=}2$: the lowest raw sliding before the legs degrade, at some cost in naturalness (\cref{sec:exp-contact}).

\begin{table*}[t]
\centering
\footnotesize
\setlength{\tabcolsep}{3pt}
\begin{tabular}{@{}l ccccccc c@{}}
\toprule
Method & Naturalness & Text & Diversity & Smoothness & Speed & Physics & Switching & Slide (cm/s)$\downarrow$ \\
\midrule
ARDY~\cite{zhao2026ardy} & 66.6 & 55.6 & 85.5 & 69.4 & \textbf{100} & \second{71.3} & 67.5 & \second{1.67} \\
DART~\cite{zhao2025dart} & 73.6 & 54.4 & 90.6 & \second{80.1} & \textbf{100} & 55.4 & 95.6 & 3.75 \\
CMDM~\cite{yu2026cmdm} & 70.8 & 61.4 & 81.5 & 78.1 & 82.3 & 61.5 & \second{96.8} & 3.33 \\
FloodDiffusion~\cite{cai2025flood} & \second{79.1} & \second{68.4} & \second{92.8} & 59.2 & 73.0 & 45.7 & 66.7 & 13.68 \\
\textbf{Ours} & \textbf{87.6} & \textbf{90.1} & \textbf{94.6} & \textbf{97.1} & \textbf{100} & \textbf{79.8} & \textbf{99.2} & \textbf{1.47} \\
\midrule
\multicolumn{9}{@{}l}{\emph{With foot post-processing}} \\
ARDY + own correction & 66.8 & \second{56.2} & 85.5 & 65.0 & -- & \second{90.4} & 63.8 & \second{0.80} \\
DART + S2-hip & \second{73.4} & 54.3 & \second{90.6} & \second{80.1} & -- & 80.4 & \second{95.4} & 1.02 \\
\textbf{Ours + S2-hip} & \textbf{87.6} & \textbf{90.0} & \textbf{94.6} & \textbf{97.2} & 100 & \textbf{96.3} & \textbf{99.4} & \textbf{0.35} \\
\bottomrule
\end{tabular}
\caption{System-level comparison in our streaming evaluation. Axes on 0--100, higher is better; speed is 100 when all real-time budgets are met; slide: median stance-ankle speed in cm/s (motion capture 0.80). CMDM and FloodDiffusion output no SMPL rotations, so S2-hip is not applied to them.}
\label{tab:main}
\end{table*}

\subsection{Training and online inference}
\label{sec:train}
\paragraph{Training stages.}
Training has three stages: the causal codec of \cref{eq:codec} is pretrained, the generator is trained on its latent (\cref{eq:gen}, 100k steps, batch 1,024), and the decoder of \cref{sec:contact} is fine-tuned on the frozen latent (\cref{fig:method}(c)).
Captions such as ``walk'' also cover turning motion (28\% of BABEL ``walk'' segments turn by over 45\textdegree), so for text that implies straight motion, we add to \cref{eq:gen} a one-sided heading loss on the decoded root yaw $\psi$,
\begin{equation}
  \mathcal{L}_\text{head}=\textstyle\sum_{W}\mathrm{Huber}_\delta\big(\big[\lvert\Delta_W\hat\psi\rvert-\max(\gamma_W,\lvert\Delta_W\psi\rvert)\big]_+\big),
  \label{eq:head}
\end{equation}
over windows $W$ of 1\,s, 3\,s, and the caption span with tolerances $\gamma_W$ of 15\textdegree, 25\textdegree, and 30\textdegree.

\paragraph{Online loop and switch-aware re-noising.}
\label{sec:switch}
At inference, each network evaluation applies \cref{eq:euler} with the guided velocity of \cref{eq:cfg}, and the oldest token, once clean, is decoded with \cref{eq:compose} into four frames and committed.
When instruction $c_j$ arrives, the tokens in flight were partly denoised under $c_{j-1}$: we roll the phase back by $r$ steps and re-noise every token in flight from its last clean prediction,
\begin{equation}
  \mathbf{z}_i\leftarrow\big(1-\tau'_i\big)\hat{\mathbf{z}}_i+\tau'_i\,\boldsymbol{\epsilon}_i,\qquad \tau'_i=\tau_i(\phi-r\,\Delta\phi),
  \label{eq:renoise}
\end{equation}
with $r{=}4$; committed frames are never touched.
In training, forcing rows are re-noised the same way with probability 0.3 and $r\sim\mathcal{U}\{1,\dots,4\}$, which smooths the transitions.
S2-hip, a causal post-process outside the model, locks a stance ankle at touchdown by rotating the hip and reduces residual sliding (\cref{eq:residual}); the robot runs both, since without it both decoders cause falls (\cref{sec:exp-robot}).

\section{Experiments}
\label{sec:exp}

\subsection{Setup}
\label{sec:exp-setup}

\begin{figure}[t]
\centering
\includegraphics[width=\columnwidth]{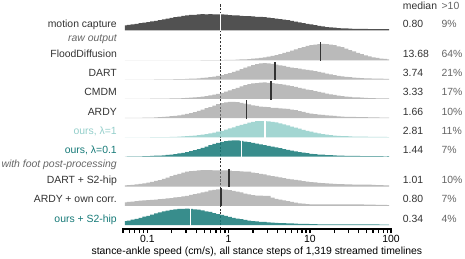}
\caption{Stance-ankle speed over all stance steps of the streamed P1 timelines; right: median (cm/s) and share above 10\,cm/s.}
\label{fig:phys}
\end{figure}

\begin{table}[t]
\centering
\footnotesize
\setlength{\tabcolsep}{1.2pt}\renewcommand{\arraystretch}{0.92}
\begin{tabular}{@{}l cccccc@{}}
\toprule
\emph{HumanML3D} & & \multicolumn{3}{c}{R-prec.$\uparrow$} \\
\cmidrule(lr){3-5}
Method & FID$\downarrow$ & @1 & @2 & @3 & MM-D$\downarrow$ & Div.$\rightarrow$ \\
\midrule
Real motion & 0 & 0.509 & 0.703 & 0.801 & 2.976 & 9.481 \\
Real, SMPL fit & 0.067 & 0.502 & 0.697 & 0.791 & 2.990 & 9.343 \\
\midrule
MDM~\cite{tevet2023mdm} & 0.582 & 0.408 & 0.596 & 0.706 & 3.653 & 9.379 \\
MLD~\cite{chen2023mld} & 0.493 & 0.459 & 0.645 & 0.749 & 3.292 & 9.552 \\
T2M-GPT~\cite{zhang2023t2mgpt} & 0.213 & 0.467 & 0.653 & 0.754 & 3.226 & 9.372 \\
MoMask~\cite{guo2024momask} & 0.091 & 0.507 & 0.700 & 0.799 & 2.975 & 9.525 \\
\midrule
MotionStreamer~\cite{xiao2025motionstreamer} & 0.406 & 0.444 & 0.633 & 0.736 & 3.396 & 9.514 \\
FloodDiffusion~\cite{cai2025flood} & 0.433 & 0.498 & 0.692 & 0.790 & 2.981 & 9.198 \\
CMDM~\cite{yu2026cmdm} & \textbf{0.146} & \textbf{0.566} & \textbf{0.768} & \textbf{0.857} & \textbf{2.648} & \textbf{9.462} \\
ARDY~\cite{zhao2026ardy} & 1.073 & 0.388 & 0.578 & 0.690 & 3.661 & 9.037 \\
\textbf{Ours} & \second{0.154} & \second{0.508} & \second{0.701} & \second{0.798} & \second{2.937} & \second{9.460} \\
\bottomrule
\end{tabular}
\caption{Comparison on HumanML3D under the standard metrics, in SMPL space; every generator except ARDY is trained only on HumanML3D. }
\label{tab:matched}
\end{table}

\paragraph{Models and data.}
\ours is trained as in \cref{sec:train} with granularity-aware conditioning and decodes with the contact-guided decoder ($\lleg{=}0.1$); \emph{+ S2-hip} adds causal foot locking.
Its training pool, the \emph{1,086\,h set} (724\,h after quality control), extends a \emph{131\,h base set} of HumanML3D~\cite{guo2022humanml3d}, BABEL~\cite{punnakkal2021babel}, and BONES-SEED with public and in-house captures, including HiPHI~\cite{ji2026hiphi}; part of the in-house takes is held out to measure coverage.

\paragraph{Streaming evaluation.}
We compare \ours with ARDY~\cite{zhao2026ardy}, DART~\cite{zhao2025dart}, CMDM~\cite{yu2026cmdm}, and FloodDiffusion~\cite{cai2025flood}, each run from its official checkpoint with its best settings.
All systems receive the same instruction streams at the same arrival times and are scored from their own joints by one TMR-style~\cite{petrovich2023tmr} evaluator trained on public data.
Single-caption segments (P0-neutral) and streamed BABEL timelines (P1) give seven axes that are reported separately: naturalness, text matching, diversity, smoothness, speed (\cref{def:stream}), physics, and switching, the last scored on the P1 switches that are detectable in ground truth.
Speed is measured on one RTX 3090 with batch size 1, foot post-processing is reported in a separate block of \cref{tab:main}, and ablations use a subset of the test sets.

\paragraph{HumanML3D benchmark.}
We also train a generator on HumanML3D only and score it with the standard evaluator~\cite{guo2022humanml3d}.
Because the robot consumes SMPL, all methods are compared in SMPL space: 263-d outputs are fitted with SMPL and converted back, as is real motion in the ``real, SMPL fit'' row of \cref{tab:matched}. Only the generator is data-matched, as the codec is pretrained on several motion-capture sets.

\subsection{Overall Streaming Performance}

\paragraph{System-level comparison.}
GC-Motion achieves the highest scores on the quality axes in Table~1 and meets the real-time budgets. Its gains include stronger text matching, smoother motion across instruction changes, and less stance-foot sliding before foot locking.

The gains extend to motion generated under successive instructions. On P1, GC-Motion achieves the lowest transition FID and the highest smoothness score. Its raw outputs also
exhibit the least stance-foot sliding, and causal foot locking reduces sliding further (Table~1 and Fig.~5). Under the switch protocol, GC-Motion responds to a greater fraction of instructions, which accounts for its higher switching score.

The systems use their respective training data. Our controlled comparisons isolate the effects of the two proposed components. With the expanded training pool held fixed, granularity-aware conditioning reduces heading drift and retains coverage of motions from the added sources (Sec.~4.3). With the generator and latents held fixed, contact-guided decoding reduces stance-foot sliding (Sec.~4.4).

\paragraph{HumanML3D comparison.}
With its generator trained on HumanML3D, GC-Motion achieves a FID close to CMDM's, while CMDM achieves higher text retrieval scores (Table~2). CMDM applies explicit motion--language supervision to its latent representation, which may contribute to its retrieval advantage. GC-Motion's codec is trained for motion reconstruction, with text conditioning learned by the generator.

\subsection{Granularity-aware conditioning}
\label{sec:exp-ablation}

\begin{table}[!tb]
\centering
\footnotesize
\setlength{\tabcolsep}{1.7pt}
\begin{tabular}{@{}l ccccc@{}}
\toprule
 & Text & Phys. & SC$\downarrow$ & Drift$\downarrow$ & Wide$\downarrow$ \\
\midrule
\multicolumn{6}{@{}l}{\emph{40k}} \\
131\,h base set & \textbf{89.1} & 82.1 & 3.95 & \textbf{8.0} & 0.182 \\
1,086\,h, naive & 86.8 & 84.6 & 1.85 & 11.8 & 0.084 \\
\oursrow 1,086\,h, granularity & 87.3 & \textbf{85.4} & \textbf{1.84} & 8.4 & \textbf{0.076} \\
\midrule
\multicolumn{6}{@{}l}{\emph{Full recipe, 100k}} \\
131\,h & 91.1 & 82.8 & 1.54 & \textbf{7.8} & 0.180 \\
1,086\,h, naive & 89.1 & 85.4 & 1.30 & 10.2 & \textbf{0.057} \\
\oursrow 1,086\,h (ours) & \textbf{91.1} & \textbf{86.2} & \textbf{1.18} & 7.9 & 0.059 \\
\bottomrule
\end{tabular}
\caption{Adding 1,086\,h of mixed-granularity data. Text and Phys.: axis scores (0--100); SC: self-collision (mm); Drift: median $|$net turn$|$ (\textdegree) on straight-walking prompts; Wide: FID on held-out takes (coverage); shaded: ours.}
\label{tab:offline}
\end{table}

\begin{figure}[t]
\centering
\includegraphics[width=\columnwidth]{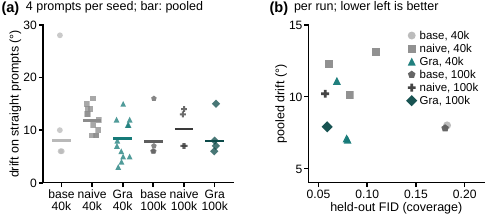}
\caption{Adding 1,086\,h of mixed-granularity data with naive or granularity-aware conditioning (Gra). (a)~Heading drift on straight-walking prompts. (b)~Drift against held-out FID; training the base set longer does not add coverage.}
\label{fig:input}
\end{figure}

\begin{figure*}[t]
\centering
\includegraphics[width=\textwidth]{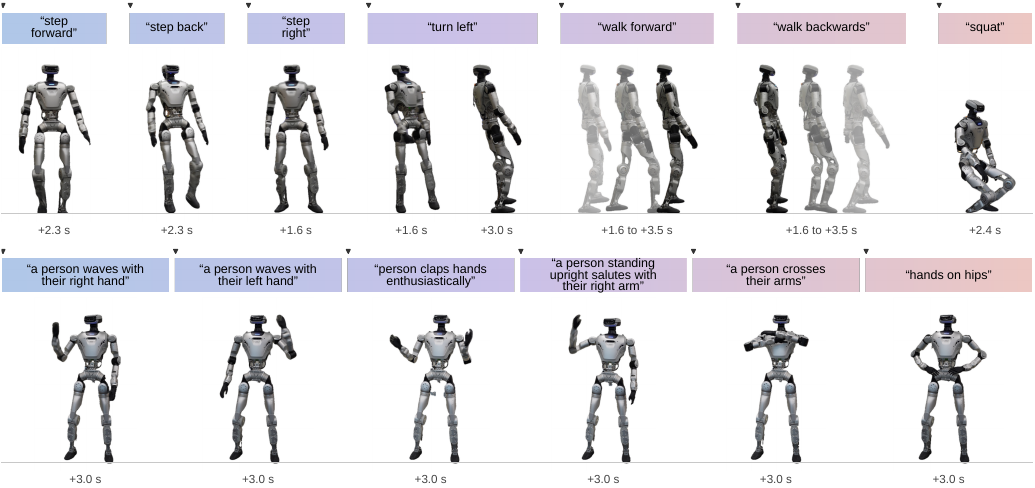}
\caption{The final model onboard the humanoid robot in live sessions. Below each instruction (arrow), frames at the given time after it, with the background removed. Top: locomotion. Bottom: upper-body instructions. More tasks are in the supplementary video.}
\label{fig:robot-stream}
\end{figure*}

\label{sec:exp-offline}
\Cref{tab:offline} separates the data from the conditioning; all rows share the heading loss of \cref{eq:head} and the caption rewriting for straight walking.
Expanding the base pool to 1,086\,h improves coverage of held-out motion from the added sources.
With the expanded pool held fixed, granularity-aware conditioning reduces heading drift on sustained forward-motion prompts while retaining this coverage gain (\cref{tab:offline,fig:input}), which supports the split of \cref{sec:offline}: aligned descriptions supervise tokens through diffusion forcing, and coarse captions supervise whole windows.
The other axes stay close to those of the base model, with switching showing the largest drop, and local retrieval after instruction switches shows no consistent improvement.

\subsection{Contact-guided decoding}
\label{sec:exp-contact}
\begin{table}[!tb]
\centering
\footnotesize
\setlength{\tabcolsep}{1.9pt}
\begin{tabular}{@{}l ccc ccccc@{}}
\toprule
 & \multicolumn{3}{c}{Round trip} & \multicolumn{5}{c}{Generation} \\
\cmidrule(lr){2-4}\cmidrule(l){5-9}
Decoder & Slide & FID & MPJPE & Slide & Phys. & +S2 & Nat. & Sw. \\
\midrule
Mocap & 0.93 & -- & -- & 0.80 & 92.5 & -- & -- & -- \\
Original & 6.70 & 0.045 & 2.33 & 5.08 & 60.5 & 86.2 & 90.7 & 94.9 \\
$1$, $0.5$ & 3.45 & 0.047 & 2.34 & 2.81 & 67.7 & 96.1 & 90.3 & 94.3 \\
$1$, $0.5$, no comp. & 8.26 & 0.053 & 2.34 & 6.56 & 58.4 & 67.2 & 90.1 & 94.7 \\
$0.1$, $0.5$ & 2.63 & 0.049 & 2.47 & 2.11 & 72.7 & 96.4 & 89.4 & 93.7 \\
\oursrow $0.1$, $2$ (ours) & 1.79 & 0.059 & 2.85 & 1.44 & 80.3 & 96.4 & 87.4 & 94.2 \\
$0$, $2$ & 1.63 & 0.067 & 3.12 & 1.31 & 81.6 & 94.9 & 86.3 & 92.8 \\
\bottomrule
\end{tabular}
\caption{Decoder ablation ($\lleg$, $\lambda_\text{p}$) with the generator fixed. +S2: physics with S2-hip; no comp.: \cref{eq:compose} off at inference; shaded: ours.}
\label{tab:contact}
\end{table}

With the generator and its latents fixed (\cref{tab:contact}), contact-guided decoding reduces stance-foot sliding in reconstruction and in generation, and raises the physics score of the generated raw output.
Disabling the composition of \cref{eq:compose} at inference makes the same trained decoder slide more than the original one, so the reduction comes from composing the root displacement with the predicted contacts rather than from the decoder's own root prediction.
\Cref{fig:committing}(b) and \cref{fig:phys} show the shift over stance phases and stance steps.
Freeing the legs (lower $\lleg$) and weighting planted feet (higher $\lambda_\text{p}$) reduce sliding further at some cost in naturalness and pose fidelity, and removing leg distillation entirely degrades the legs.
With causal foot locking on top, every contact-guided decoder reaches higher physics than the original decoder, so the two are complementary.

\subsection{Humanoid deployment}
\label{sec:exp-robot}
We deploy the complete system with causal foot locking on a LimX Oli humanoid~\cite{limx_oli} (\cref{fig:teaser}(b)): a C++ node on its Jetson Orin NX generates motion onboard and feeds a kinematic retargeter and the whole-body tracking controller of COSA~\cite{fu2026cosa05}, and generation commits within the per-token budget in steady state.
\Cref{fig:robot-stream} shows frames from live sessions in which the robot follows typed instructions one after another; the supplementary video contains the complete sessions.
In MuJoCo with the same node, retargeter, and controller, foot locking prevents the falls that both decoders cause without it; with it, contact-guided decoding lowers stance slip, tracking error, and peak tilt relative to the original decoder, at the cost of longer half squats.
On the real robot, single instructions and scripted sequences ran without an observed fall.

\section{Conclusion}
\label{sec:conclusion}
In \ours, the time span of a caption decides how it supervises a streaming generator: aligned text conditions tokens and coarse captions whole windows, so 1,086\,h of mixed-granularity capture add coverage while heading drift stays near that of the base set. Computing the root from predicted contacts lowers the raw stance-ankle speed from 6.4 to 1.8 times that of motion capture, and with causal foot locking (S2-hip) the system runs onboard a humanoid robot in real time.
\emph{Limitations:} inferred boundaries are uncertain; the added data broaden coverage more than text matching; and lower sliding costs some naturalness and pose fidelity. Future work includes aligning the latent with text, which may narrow the retrieval gap on HumanML3D, and moving foot locking into the decoder so that the robot no longer needs a post-process.

{
    \small
    \bibliographystyle{ieeenat_fullname}
    \bibliography{main}
}

\end{document}